\documentclass[letterpaper, 10 pt, journal, twoside]{ieeetran}

\IEEEoverridecommandlockouts                              

\usepackage[utf8]{inputenc}
\usepackage{url}
\usepackage{graphics} 
\usepackage{graphicx} 
\usepackage{mathptmx} 
\usepackage{times} 
\usepackage{amsmath} 
\usepackage{amssymb}  
\usepackage{color}
\usepackage{xcolor}
\usepackage{subfig}
\usepackage{epstopdf}
\usepackage{balance}
\usepackage{wrapfig}
\usepackage{adjustbox}
\usepackage{booktabs, makecell}
\usepackage{multirow}
\usepackage{times}
\usepackage{textcomp}
\usepackage{todonotes}
\usepackage{dsfont}
\usepackage{array}
\usepackage{tensor}
\usepackage{microtype}

\usepackage{cite}

\usepackage{algpseudocode}
\usepackage[ruled,linesnumbered]{algorithm2e}

\renewcommand{\baselinestretch}{0.999}
    
\DeclareMathAlphabet{\mathcal}{OMS}{cmsy}{m}{n}

\def\R{\ensuremath{\mathbb{R}}}

\def\A{\ensuremath{\mathcal{A}}}
\def\D{\ensuremath{\mathcal{D}}}
\def\Cset{\ensuremath{\mathcal{C}}}

\def\rle{\ensuremath{\prec}}

\newcommand{\var}[1]{\ensuremath{V\left({#1}\right)}}

\newcommand{\app}[2]{\ensuremath{{#1}(#2)}}

\renewcommand{\vec}[1]{\ensuremath{\mathbf{#1}}}
\newcommand{\mat}[1]{\ensuremath{\mathbf{#1}}}
\newcommand{\norm}[1]{\left\lVert#1\right\rVert}
\newcommand{\fd}[1]{\ensuremath{\dot{#1}}}
\newcommand{\sd}[1]{\ensuremath{\ddot{#1}}}

\newcommand{\diff}[2]{\ensuremath{\frac{\partial#1}{\partial#2}}}

\newcommand{\vecM}[1]{\ensuremath{\begin{bmatrix} #1 \end{bmatrix}}}
\newcommand{\set}[1]{\ensuremath{\left\{#1\right\}}}
\newcommand{\fset}[2]{\ensuremath{\set{#1 \;\middle\vert\; #2}}}

\newcommand{\ce}[2]{\ensuremath{{#1}_{#2}}}
\newcommand{\lb}[1]{\ensuremath{\ce{\iota}{#1}}}
\newcommand{\ub}[1]{\ensuremath{\ce{\upsilon}{#1}}}

\newcommand{\tf}[3]{\tensor[^{#1}]{\mat{#2}}{_{#3}}}
\newcommand{\Csset}[1]{\ensuremath{\Cset \left(#1 \right)}}

\newcommand{\code}[1]{\texttt{#1}}

\newcommand{\eg}{\mbox{e.\,g.}\xspace}
\newcommand{\ie}{\mbox{i.\,e.}\xspace}

\newcommand{\etal}{\emph{et al.}\xspace}
   
\renewcommand{\[}{\begin{equation}}
\renewcommand{\]}{\end{equation}}

\newcommand{\secref}[1]{Sec.~\ref{#1}}
\renewcommand{\eqref}[1]{Eq.~(\ref{#1})}
\newcommand{\figref}[1]{Fig.~\ref{#1}}
\newcommand{\tabref}[1]{Tab.~\ref{#1}}

\newcommand{\projectsite}{\url{http://kineverse.cs.uni-freiburg.de}}
\newcommand{\sectitle}[1]{\textbf{#1:}}
\renewcommand{\baselinestretch}{0.99}

\long\def\ignore#1{}

\def\bedit{\color{black}}
\def\eedit{\color{black}}
\newcommand{\sedit}[1]{\bedit #1 \eedit}

\usepackage[hidelinks]{hyperref}

\usepackage{xcolor}
\usepackage{mdframed}
\usepackage[shortlabels]{enumitem}

\newlength\myindent
\title{
Kineverse: A Symbolic Articulation Model Framework for Model-Agnostic Mobile Manipulation
}

\author{Adrian Röfer$^{1,2}$, Georg Bartels$^2$, Wolfram Burgard$^1$, Abhinav Valada$^1$, and Michael Beetz$^2$
\thanks{Manuscript received: September, 10, 2021; Revised December, 9, 2021; Accepted January, 5, 2022.}
\thanks{This paper was recommended for publication by Editor Lucia Pallottino upon evaluation of the Associate Editor and Reviewers' comments.
This work was funded by the BrainLinks-BrainTools center
of the University of Freiburg.}
\thanks{$^1$ Department of Computer Science, University of Freiburg, Germany.}
\thanks{$^2$ Institute for Artificial Intelligence, University of Bremen, Germany.}
\thanks{Digital Object Identifier (DOI): see top of this page.}
\thanks{© 2022 IEEE.  Personal use of this material is permitted.  Permission from IEEE must be obtained for all other uses, in any current or future media, including reprinting/republishing this material for advertising or promotional purposes, creating new collective works, for resale or redistribution to servers or lists, or reuse of any copyrighted component of this work in other works.}
}

\begin{document}

\maketitle

\begin{abstract}
Service robots in the future need to execute abstract instructions such as ''fetch the milk from the fridge''. To translate such instructions into actionable plans, robots require in-depth background knowledge.
With regards to interactions with doors and drawers, robots require articulation models that they can use for state estimation and motion planning.
Existing frameworks \bedit model articulated connections as abstract concepts such as \emph{prismatic}, or \emph{revolute}, but do not provide a parameterized model of these connections for computation. \eedit
In this paper, we introduce a novel framework that uses symbolic mathematical expressions to model articulated \bedit structures -- robots and objects alike -- in a unified and extensible manner\eedit.
We provide a theoretical description of this framework, and the operations that are supported by its models, and introduce an architecture to exchange our models in robotic applications, making them as flexible as any other environmental observation.
To demonstrate the utility of our approach, we employ our practical implementation \emph{Kineverse} for solving common robotics tasks from state estimation and mobile manipulation, and use it further in real-world mobile robot manipulation.
\end{abstract}



\section{Introduction}
\label{sec:introduction}

\IEEEPARstart{F}{or} robots to become ubiquitous helpers in our everyday lives, they will need to be \sedit{instructable} or programmable with abstract, high-level commands. During a cooking session, a robot might be instructed to \emph{Fetch the milk from the fridge}, \emph{Get the flour out of the pantry}, or \emph{Bring me a bowl}.
Faced with such an abstract instruction, \bedit the robot needs to derive an actionable plan. This requires reasoning over and generation of possible interactions in the environment. Ideally, these capabilities should be universal across different robots and objects so that new robots can be deployed easily and novel objects can be interacted with. Many objects in everyday environments such as doors, drawers, desk lamps, knobs, switches, or ironing boards are, much like robots themselves, a collection of smaller objects which are connected by mechanical linkages and have a reduced range of motion. \eedit
This class of objects is commonly referred to \emph{articulated objects}.

\begin{figure}[t!]
\centering
\includegraphics[width=\linewidth]{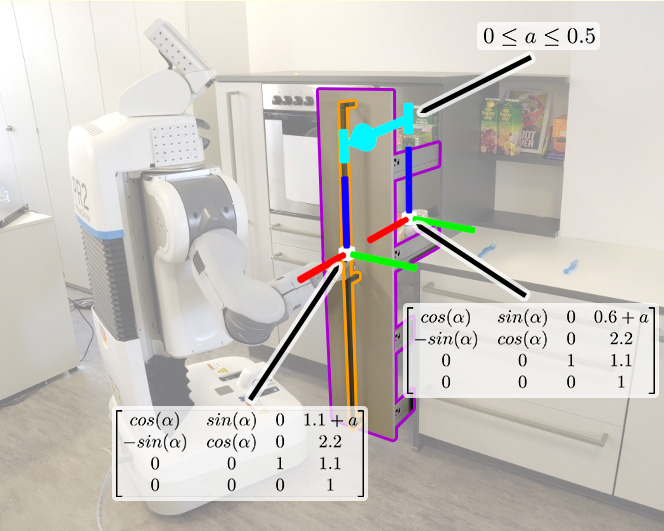}
\caption{Symbolic representation of the forward kinematics of a drawer and its handle, including positional limits..}
\label{fig:teaser}
\end{figure}

\bedit We see two challenges with respect to modeling articulated structures in robotics today. Existing frameworks for this purpose model these structures as a series of rigid bodies which are connected by different types of joints. The first challenge arises from existing frameworks not providing a mathematical model for the joint types, which is needed for manipulations as depicted in \figref{fig:teaser}. They describe connections as \emph{prismatic}, or \emph{revolving}, but do not define how these connections are translated to relative transformations between bodies, requiring every user of the models to create their own definitions. This might not be a problem where the set of joint types is sufficient for modeling all objects, however, objects such as garage doors, and robotic actuation via tendons cannot be modeled with the existing frameworks. 

The second challenge we see is with respect to the flexibility of articulated objects within an active robotic system. When a robot picks up an object, this object can now be viewed as an additional body part of the robot. Similarly, as a robot enters a new environment, it has to hypothesize about the articulations of external objects around it in order to interact with them. In both cases, state estimation, and manipulation components need to have access to the same articulation model of the robot or external objects in order to plan and execute robotic actions in the environment. Currently, there are standards for exchanging camera images and even spatial transformations within robotic applications, however, none of these exist for models of articulated structures, since these are viewed as static even though they actually are not. \eedit

With this paper, we seek to address both challenges. 
\bedit Our goal is to introduce an articulation model framework that can be used to model articulated structures without using fixed types of articulation. The generated models can directly be used to perform forward kinematic computations without requiring any knowledge about the types of articulations such as \emph{prismatic}, or \emph{revolving} connections, which will enable our framework to accommodate any articulation. Our models are to be fully analytically differentiable, as this is a common requirement for many methods in robotics. We also want to consider the practical implications of such models and thus aim to propose a method of managing and utilizing them in a modern robotics application. 
We hold that such a computationally complete and extensible framework for articulated structures can overcome the aforementioned challenges and enable the development of articulation-agnostic robotic skills -- skills that transfer seamlessly between different articulated structures, be these robots, or external objects. \eedit

Our main contributions in this paper are:
\begin{enumerate}[topsep=0pt]
\item We introduce an approach to represent articulated \sedit{structures} as a set of constrained, symbolic mathematical expressions and a method of constructing \sedit{respective} models using this representation. 
\item \bedit We provide a reference implementation of our proposed framework, \emph{Kineverse}, through which we address the runtime aspect of employing our proposed approach. \eedit
\item We exemplify how our approach connects with robotics problems, by presenting a number of simulated and real-world experiments. \bedit We demonstrate that our models enable the development of model-agnostic robotic skills. \eedit
\end{enumerate}

The implementation of our framework and videos of all the experiments can be found at \projectsite.

\section{Related Work}
\label{sec:related_work}

In robotics, there are a few frameworks for modeling kinematics and articulated \bedit structures\eedit. Within the ROS~\cite{quigley2009ros} ecosystem, URDF and SDF are the most notable ones. Both are XML-based and partition articulated \sedit{structures} into their rigid bodies (\emph{links}) and the connections between them (\emph{joints}).
The standards define a number of joint types such as \emph{fixed}, \emph{revolute}, \emph{prismatic}, and \emph{planar}. A joint instance can be parameterized, \eg the axis of rotation for rotational joints can be specified. The configuration space of each joint can be limited in position and velocity. SDF expands on URDF in joint types and adds the concept of \emph{frames}.

COLLADA~\cite{arnaud2006collada} is a format common to the world of computer graphics. It is capable of encoding articulation models and provides more flexibility than URDF and SDF. Instead of providing a fixed set of joint types, it models joints by limiting the number of degrees-of-freedom (DoF) of a 6-DoF connection. It allows developers to define expressions for many of the parameters within a model file, enabling the creation of joints that mimic others, and to create state-dependent joint limits. 

All these frameworks take a descriptive approach to modeling articulated \bedit structures\eedit, describing articulated structures using higher-level constructs such as \emph{hinged}, or \emph{prismatic} connections. This abstraction makes the models very approachable to human developers. However, currently, the models do not encode rules for constructing the underlying \sedit{mathematical} representation of the articulated \bedit structures. Generally, these frameworks encode kinematic chains, the forward kinematic expressions for which can be obtained by multiplying the relative transformations of their parts, but they do not specify an exact mathematical representation for the modeled articulations. For the simpler types, it is possible to deduct what the mathematical model is supposed to be, for more complicated types, such as \emph{planar} and \emph{floating} in the case of URDF, no detailed description of these joints is given. Further, due to the lack of parameterization, no way of constraining these articulations is provided. A similar problem can be found in SDF with the \emph{ball} and \emph{universal} articulation types. Even if the documentation of these frameworks stated clear mathematical models for their modes of articulation, human intervention would still be required to implement them in software for use in computation. This effort would be required over and over again on the part of thousands of entities if a new mode of articulation was introduced to these frameworks, as the frameworks themselves cannot transport the mathematical model of the articulations. \eedit

As we strive to create a new articulation model framework that supports direct computation, we must be mindful of common computational requirements made by today's robotic applications.
We see three common model applications: Model estimation, state estimation, and motion generation.

\bedit To collect models of objects that can be stored using frameworks such as ours, works such as~\cite{sturm2011probabilistic, ruhr2012generalized, martin2014online, hausman2015active} have addressed the problem of model estimation\eedit. All of them detect rigid body motions from image sequences and assess the probability of the motion of an object pair to be explained by a particular type of connection. This assessment relies on an inverse kinematic projection into the configuration space of an assumed connection and then compares the forward kinematic projection to the actual observation.
While the earliest approach in this line of works~\cite{katz2008manipulating} does not recover the full kinematic structure of the object, but only the likeliest pair-wise connections, later work \cite{sturm2011probabilistic, baum2017opening} recover the full kinematic graph of the model.
To generate observations for an estimation, \cite{ruhr2012generalized, barragan2014interactive, hausman2015active, baum2017opening} \bedit use simple heuristics for interactions\eedit.
Instead of such exploratory behaviors, an initial model can also be generated using trained models~\cite{hofer2016coupled, li2020category} and then verified interactively.

To be of use for robotic manipulation, a new articulation model framework needs to meet the requirements of existing motion generation methods. For Cartesian Equilibrium Point control \cite{jain2010pulling} as it is used in \cite{sturm2011probabilistic, ruhr2012generalized, baum2017opening} only forward kinematics are required. The same is also a requirement for learning-based control approaches such as \cite{honerkamp2021learning}. 
Sampling-based trajectory planners such as RRT and PRM~\cite{lavalle2006planning} require configuration space bounds to determine the space to sample from.
Optimization-based approaches, be it planners~\cite{ratliff2009chomp, schulman2013finding, toussaint2015logic} or control~\cite{aertbelien2014etasl, fang2016learning}, require model differentiability. Our approach \sedit{satisfies} the \sedit{technical} requirements for forward kinematic expressions, model differentiability, and modeling constraints on systems of equations and inequations \bedit made by these existing approaches\eedit.

\section{Technical Approach}
\label{sec:methods}

In this section, we introduce our approach for building flexible, and scalable models of articulated \sedit{structures} using symbolic mathematical expressions. \bedit In the second part of this section, we describe our reference implementation of the proposed model and its integration with existing robotic systems. \eedit

\subsection{Articulation Model}
\label{sec:articulation_model}

We propose modeling a scene of articulated \sedit{structures, consisting of both robots and objects}, as a tuple $\A = (\D_{\A}, \Cset_{\A})$, where $\D_{\A}$ is a named set of forward kinematic (FK) expressions of different \sedit{structures} and $\Cset_{\A}$ is a set of constraints which restricts the validity of parameter assignments for the expressions in $\D_{\A}$. In the following sections, we introduce the elements of these two collections separately.

\subsubsection{Expressions in $\D_{\A}$}

We base our approach on symbolic mathematical expressions. The expressions $\varphi \in \D_{\A}$ can be either singular expressions, evaluating to $\R$, or matrices evaluating to $\R^{m\times n}$. As a convention, we use Greek letters $\varphi, \psi$ to refer to singular expressions, bold letters $\vec{u}, \vec{v}$ to refer to \bedit vector typed expressions\eedit, and bold capital letters $\mat{M}, \mat{T}$ to refer to \bedit matrix typed expressions\eedit. We use these expressions to encode the FK of rigid bodies in articulated objects as $\R^{4\times 4}$ homogeneous transformations. \eqref{eq:ex_transform} shows an example of a transformation $\mat{T}$ translating by $b$ meters along the z-axis and rotating by $a$ radians about the same.\looseness=-1
\[
\label{eq:ex_transform}
\mat{T} = \vecM{ \cos a & \sin a & 0 & 0 \\
				-\sin a & \cos a & 0 & 0 \\
				      0 &      0 & 1 & b \\
				      0 &      0 & 0 & 1}
\]

As defined in \eqref{eq:ex_transform}, $\mat{T}$ is parameterized under $a, b$. We call the set of these parameters the set of \emph{variables} of $\mat{T}$ and refer to it by $\var{\mat{T}} = \set{a, b}$. These variables can be substituted with real values to evaluate the value of $\mat{T}$ at a specific \emph{configuration space position}. Borrowing the notation convention for configuration space positions from robotics, we specify these positions as substitution mappings $\vec{q} = \set{q_1 \rightarrow x_1, \ldots, q_n \rightarrow x_n}$ mapping variables to values in $x_i \in \R$.
Applying such a mapping $\vec{q}_1 = \set{a \rightarrow \pi, b \rightarrow 2}$ to $\mat{T}$ would evaluate $\mat{T}$ at this position:
\[
\app{\mat{T}}{\vec{q}_1} = \vecM{ -1 &  0 & 0 & 0 \\
								   0 & -1 & 0 & 0 \\
								   0 &  0 & 1 & 2 \\
								   0 &  0 & 0 & 1}
\] 

\bedit The variables model the degrees of freedom of our articulated structures. As it is important to be able to address not only the position of a DoF but also its velocity, acceleration, etc., \eg to constrain the maximum velocity of a robotics joint, we introduce the following convention: For a degree of freedom $a$, $a$ refers to its position, $\fd{a}$ to its velocity, $\sd{a}$ to its acceleration, etc. \eedit
Symbolic mathematical frameworks are able to generate analytical derivatives for an expression, \ie given $\varphi = \sin a + b^2$ we can generate the derivative expressions for all $v \in \var{\varphi}$ as
\[
\label{eq:ex_diff}
\begin{aligned}
&\diff{\varphi}{a} = \cos a, & \diff{\varphi}{b} = 2b.
\end{aligned}
\]

In the same way it is also possible to construct the analytical gradient $\nabla \varphi$ w.r.t all $v \in \var{\varphi}$:
\[
\nabla \varphi = \vecM{\cos a, 2b}^T
\]

Under our approach, we change the definition of an expression's gradient from a column vector of derivatives to a mapping. This mapping associates the derivative $\fd{v}$ of a variable $v$ with the derivation of \sedit{$\varphi$} w.r.t $v$, \ie
\[
\nabla \varphi := \set{\fd{a} \rightarrow \cos a, \fd{b} \rightarrow 2b }.
\]

Aside from maintaining semantic information about the gradient, we introduce this redefinition in order to accommodate two special modeling problems in the field of robotics: Non-holonomic kinematics, and the definition of heuristic gradients.
In the case of non-holonomic kinematics, such as the differential drives common to many robots, \bedit there is no function that allows us to determine the position of the robot in the world from the joint space position of its wheel, yet there is a function mapping the wheels' velocities to a velocity of the robot's base. To address this problem \eedit we allow the gradient to include components for \sedit{$\diff{\varphi}{v}$} where \bedit $v \not \in \var{\varphi}$\eedit. Under this relaxation, the following represents a valid gradient for $\varphi$:
\[
\label{eq:ex_expanded_grad}
\nabla \varphi = \set{\fd{a} \rightarrow \cos a, \fd{b} \rightarrow 2b, \underline{\fd{c} \rightarrow 4, \fd{d} \rightarrow -4} }
\]

\bedit Referring back to differential drives: $\fd{c}, \fd{d}$ would represent the velocity of the left and right wheel.\eedit

A second modeling challenge we seek to accommodate is modeling kinematic switches. Baum~\etal~\cite{baum2017opening} examined the problem of exploring lockboxes, articulated systems in which the movability of one part is dependent on the state of another part. While lockboxes might be an exotic example, the same principle applies to everyday objects such as doors. We want to model such mechanisms with our approach\bedit. While Baum \etal use a model based on explicit change points and zones throughout which the DoF remains locked or unlocked, first introduced by~\cite{kulick2015active}, we opt to use a continuous model to represent such states by using high contrast sigmoid functions. As these functions' derivatives outside of the region of change are effectively $0$, we introduce a further extension to our gradient formulation: We permit the overriding of derivatives the gradient as\eedit
\[
\nabla \varphi = \set{\fd{a} \rightarrow \cos a, \underline{\fd{b} \rightarrow 1}, \fd{c} \rightarrow 4}.
\]

\bedit We aim to enable searching the configuration space of our models heuristically using these gradients. \eedit
Though these gradients are not derived from the original expression, they are propagated in accordance with the laws of differentiation, an example for multiplication can be
\[
\label{eq:ex_propagation}
\begin{aligned}
\psi &= 4a \\
\nabla \varphi \psi &= \set{\fd{a} \rightarrow 4(\sin a + a \cos a + b^2), \fd{b} \rightarrow 4a, \fd{c} \rightarrow 16a}.
\end{aligned} 
\]

Note that the non-overwritten derivative of $\diff{\varphi\psi}{b}$ would be $\diff{\varphi\psi}{b} = 8ab$. \\

\subsubsection{Constraints in $\Cset_{\A}$}
\label{sec:constraints}

We introduced the components of $\D_{\A}$ in the previous section. In this section, we describe their counterparts, the constraints, contained in $\Cset_{\A}$. While the components of $\D_{\A}$ describe the kinematics of \bedit articulated structures\eedit, the constraints $c \in \Cset_{\A}$ are used to determine which value assignment $\vec{q}$ is a valid assignment for the model $\A$. 
The constraints $c \in \Cset_{\A}$ are three-tuples $c = (\lb{c}, \ub{c}, \ce{\varphi}{c})$ which model a dual inequality constraint $\lb{c} \leq \ce{\varphi}{c} \leq \ub{c}$. This dual inequality can be used to express ranges as well as single inequalities and equalities. The \sedit{lower and upper bounds}  $\lb{c}, \ub{c}$ can be either constants or expressions themselves, \bedit enabling state dependent constraints\eedit. However, $\ce{\varphi}{c}$ is always expected to be an expression. Based on this definition of constraints, we define the subset of \emph{relevant} constraints for an expression $\psi$ \bedit on the basis of the variables it shares with all constraints' constrained expressions $\ce{\varphi}{c}$ as \eedit $\Csset{\psi} = \fset{c}{\var{\ce{\varphi}{c}} \cap \var{\psi} \not = \emptyset}$.  \bedit If the constrained expression and $\psi$ have variables in common, the constraint affects the valid variable assignments $\vec{q}$ for $\psi$.\eedit

Articulation model frameworks such as URDF offer the ability to constrain different aspects of a model's DoF such as its position, velocity, etc. In our approach, this is expressed by the variables used in the constrained expression $\ce{\varphi}{c}$ of a constraint. 
Given \bedit our convention for associating variables with DoF, \eedit 
$c_1$ in \eqref{eq:ex_constraint} limits the position of DoF $b$, while $c_2$ limits its velocity.
\[
\label{eq:ex_constraint}
\begin{aligned}
&c_1 = (-1, 2, b), &c_2 = (-1, 2, \fd{b})
\end{aligned}
\]

\subsubsection{Model Building}

In existing \sedit{frameworks} articulated \sedit{structures} consist of a set of rigid links which are connected pairwise using joints. \sedit{Formally, the} graph of connections is required to be a tree\bedit, so it should be \eedit simple to construct the forward kinematic transformation of a link, given a way of translating the connection types to mathematical objects by simply walking up and down the branches of the tree. \bedit However, frameworks such as URDF allow joints to mimic other joints on the same branch, even when they are further from the root, thus actually creating a directed acyclic graph. We \eedit do not want to \bedit introduce a complicated procedure for parsing our models, which is why \eedit
we propose a more general approach to model building. We view model building as a sequential process in which \emph{operations} are applied to expand and change a model. These operations are functions $o : X^n \Rightarrow \D\times\Cset$ which produce updates for the sets $\D_{\A}, \Cset_{\A}$ of a model, given a number of arguments $x_i \in X$. These arguments $x_1,\ldots,x_n$ can be either values from $\D_{\A, t}$ or constant values. Under this view, an articulation model is the product of the application of operations $o_1(\vec{x}_1), \ldots, o_n(\vec{x}_n)$ to an initial empty model $\A_0 = (\emptyset, \emptyset)$:
\[
\A_n = \A_0 \vDash_{o_1(\vec{x}_1)} \A_1 \vDash_{o_2(\vec{x}_2)} \A_2 \ldots \vDash_{o_n(\vec{x}_n)} \A_n 
\]

Aside from storing the order in which these operations are applied, we also propose associating them with semantic \emph{tags} such as \code{connect A B}, where \code{A},\code{B} are frames stored in the model. Using these tags it is possible to determine where an additional operation modifying \code{A} should be inserted into the sequence if it is to affect \code{B}. In relation to the task of estimating models of articulated objects from observations, \bedit the operations should represent self-contained steps in building a model, such as adding a body to the model, and, in a second operation, forming a connection with another body. \eedit
\bedit
\subsection{Reference Implementation}

As part of our proposal, we provide a reference implementation of the framework we proposed in the previous sections. The model is implemented as a Python package. We use Casadi~\cite{Andersson2019} as the backbone for our symbolic mathematical expressions. Using this backbone, we add one additional type for expressions and matrices with extended gradients, as per our proposed model. Operator overloading and function overriding add the required propagation of gradients when combining these objects with other expressions.

We also implement a base class for the operational mechanism we propose and use it to implement operations realizing the types of articulations necessary for our experiments. To be compatible with existing URDF models, we derive operations for all articulations supported by URDF and add a loader that creates a sequence of these operations from a URDF file. The operations first add bodies to the model's set $\D_\A$, including their relative transformations to their parent joints as described in the URDF. After all bodies have been created, the remaining operations form the connections by instantiating a parameterized relative transformation for each joint and multiplying these with the transformations of the parent and child bodies, updating the contents of $\D_{\A}$. The operations forming the connections also add their respective joint's configuration space constraints to the model's constraints $\Cset_{\A}$.

Our implementation also contains a number of ROS-specific bindings which enable the exchange of articulation models among the components of an active robotic system. With these bindings, we satisfy the requirement we stated in \eedit \secref{sec:introduction} for these models to be exchangeable and updatable in a live robotic system. \bedit
We realize this requirement as it is important for understanding the impact of our proposed approach on an active robotic system.
In our implementation, we opt for a simple client-server architecture in which a central model server maintains the articulation model for the running robotic system. Clients can subscribe to the model, or parts of it, which results in an on-change hook being invoked when the part of interest is changed. Clients can also introduce changes to the model stored by the server, which results in an update in all subscribers of the changed parts. For greater detail on this aspect of our approach, we refer you to our project page.

As geometric queries are highly important for motion generation, \eg in collision detection, our implementation also includes an integration with the Bullet physics engine~\cite{coumans2021}. Using this integration, collision objects affected by symbolic DoF can be identified, and queries for distances and contact normals between objects can be formulated and evaluated. We use this feature in the later parts of our evaluation. 
\eedit

\section{Experimental Evaluation}
\label{sec:experiments} 

In this section, we evaluate our proposed framework. First, we model a conditional kinematic which URDF and SDF cannot represent, to demonstrate that our approach has greater modeling capabilities.
Second, we evaluate the applicability of our framework for mobile manipulation by designing model agnostic controllers to interact with articulated objects.
Lastly, we demonstrate the utility of our framework for real-world mobile manipulation by deploying the controllers on a real robotic system. 
Throughout this evaluation, we use our practical implementation of the concepts, \emph{Kineverse}, introduced in \secref{sec:methods}. 
\bedit The \eedit code for all the experiments, and videos of all experimental results can be found at \projectsite. 

\subsection{Modeling a Novel \sedit{Mode of Articulation}}

Sturm~\etal~\cite{sturm2011probabilistic} used Gaussian processes to model the kinematics of articulated objects which cannot be expressed as a hinged or prismatic connection. One of their observed objects is a garage door which has a simple 1 DoF kinematic and falls into this category. Neither URDF nor SDF can model this mechanism.
We model this \sedit{mode of articulation} to demonstrate the greater modeling capabilities of our approach.

\begin{figure}
\setlength{\fboxrule}{0.0pt}
\framebox[8.5cm]{}
\centering
\includegraphics[width=7cm]{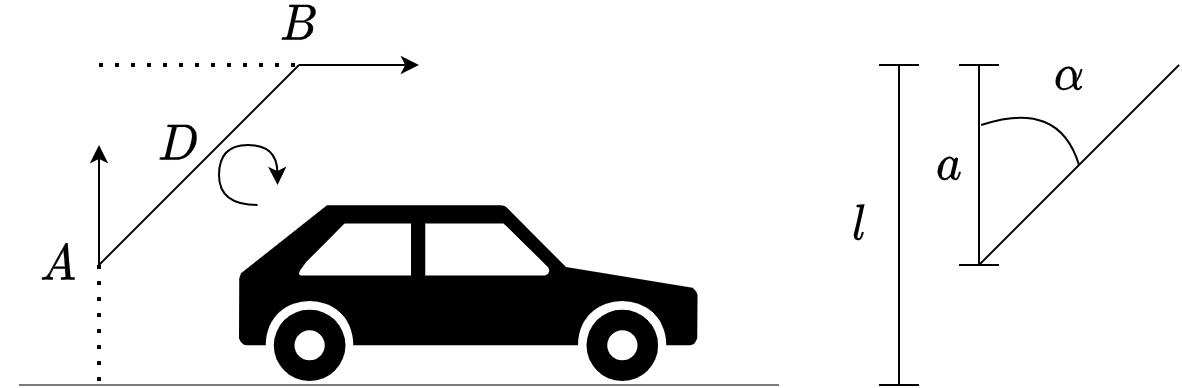}
\caption{Schematic depiction of the garage door kinematic. Left: A garage door $D$ is mounted on two hinges $A, B$ which slide along two rails, leading to a simultaneous linear motion and rotation of the door. Right: The mechanism reduced to a constant rail length $l$, a single Dof $a$, and the angle of door rotation $\alpha$.}
\label{fig:ex_garage_door}
\end{figure}

As the subject for this exercise, we pick a rigid garage door as depicted in \figref{fig:ex_garage_door}. The door $D$ is attached to two hinges $A, B$ which slide along two rails, one vertical and one horizontal.
We opt to model the door's transformation $\tf{W}{T}{D}$ as a rotation around $A$ which in turn translates along the Z-axis as
\[
\tf{W}{T}{D} = \tf{W}{T}{A} \cdot \tf{A}{R}{D}.
\]

We parameterize the state of the door using position $a \in [0, l]$. At position $a = 0$ the door is fully open and at $a = l$ the door is fully closed. Given this parameterization, we can view the problem of finding the matching angle of rotation $\alpha$ for the door as $\alpha = \cos^{-1}\frac{a}{l}$.
For our specific door, we choose $l = 2$ and arrive at the following forward kinematic transformation for the door:
\[
\begin{aligned}
\tf{W}{T}{D} &= \tf{W}{T}{A} \cdot \tf{A}{R}{D} \\
             &= \vecM{ \frac{a}{2} & 0 & \sqrt{1 - \frac{a}{2}^2} & 0 \\
                                 0 & 1 &                        0 & 0 \\
                      -\sqrt{1 - \frac{a}{2}^2} & 0 & \frac{a}{2} & a \\
                                              0 & 0 &           0 & 1 }
\end{aligned}
\]

In addition to our primary DoF $a$, we add a second DoF $b$ which serves as a lock for the door. For the sake of brevity, we do not add another transform for this locking mechanism to our model. We constrain $a$ using $c_1 = (0, 2, a)$ and $\fd{a}$ using $c_2 = (-unlocked, unlocked, \fd{a})$, where $unlocked = 1 - (b \rle 0.3) \cdot (1.99 \rle a)$. 
The expression \sedit{evaluates to} $0$ when the door is (almost) fully closed \sedit{at $a=2$} and $b$ is not at least at position $0.3$ or higher\bedit, otherwise the expression evaluates to $1$. By referencing $a$ this model mimics that the door needs to be at the right position for the bolt to be able to lock it\eedit. Our complete model is:
\[
\label{eq:garage_model}
\begin{aligned}
\A_{door} &= (\D_{door}, \Cset_{door}) \\
\D_{door} &= \{\text{door\_in\_world} : \tf{W}{T}{D}\} \\
\Cset_{door} &= \{c_1 : (0, 2, a), \\
			 &	 c_2 : (-unlocked, unlocked, \fd{a})\}
\end{aligned}
\]

\begin{figure}[t]
    \setlength{\fboxrule}{0.0pt}
    \framebox[8.5cm]{}
    \centering
    \includegraphics[height=3.8cm]{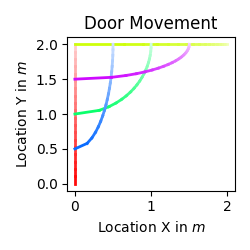}
    \includegraphics[height=3.8cm]{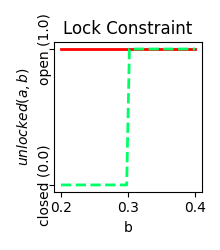}
    \caption{Plots of the garage door model. Left: Movement of $A$(red), $B$(lime), and of points $25\%$ (blue), $50\%$ (green), and $75\%$ (purple) up the door; Right: Plot $1-locked$ at $a=2$ (green) and $a=1$ (red).}
    \label{fig:garage_plots}
\end{figure}

We conclude our modeling evaluation by plotting the paths of $A, B$, and of three points on the door. In addition, we plot $\lb{c_2}$ over $b$ at $a = 2$ and $a = 1$ to demonstrate that our model can accommodate such binary locking constraints. The plots are shown in \figref{fig:garage_plots}. The model is able to produce the paths of a garage door while the locking constraint constrains $\fd{a}$ only when $b$ is in a locking position, and $a$ is in a position where it can be locked. Thus, our approach accommodates novel \sedit{mode of articulation} and binary switches.

\subsection{Model-Agnostic Object Tracking}

In this section, we use our approach in a state estimation task for articulated object manipulation. We use an Extended Kalman Filter (EKF) as a state estimation technique.
We evaluate our framework by using it to estimate the configuration space pose of the IAI-kitchen\footnote{IAI kitchen: \url{https://github.com/code-iai/iai_maps}} from observations of the 6D-poses of its parts. The aim of this experiment is to demonstrate how our approach connects with such a task and to assess the performance of our approach in it.

\sedit{We load the model of the kitchen from a URDF file into our model framework. The EKF only has access to the differentiable forward kinematics of the kitchen and can query relevant constraints based on variables its forward kinematic expressions.}

\sectitle{EKF Model Definition}
To estimate the pose $\vec{q}$ of an articulated object, our EKF consumes a stream of observations $Y_t \subseteq SE(3)$ of the object's individual parts, as they might be obtained from tracking markers~\cite{jurado2015generation} or pose estimation networks~\cite{xiang2018posecnn}.
We use the forward kinematics $h(\vec{q})$ provided by our model $\A$ for the measurement prediction $h(\vec{q})$. We assume normally distributed noise in the translations and rotations of the parts' poses. Assuming a variance of $\sigma_t, \sigma_r$ in translational and rotational noise respectively, we generate the covariance $\mat{R}$ of our observations by sampling observations with added noise using $h$.
We exploit the symbolic nature of our framework to reduce the $\R^{4\times 4}$ observation space of this task by selecting only matrix components $y_{ij}$ which are symbolic, \ie $\var{y_{ij}} \not = \emptyset$.

We define a simple prediction model $f(\vec{q}, \fd{\vec{q}}) = \vec{q} + \Delta t \fd{\vec{q}}$ where the control input is a velocity in the object's configuration space. In this experiment, we assume the object to be completely static and thus set the prediction covariance $\mat{Q} = \mat{0}$.
Note that both $f, h$ satisfy the EKF's requirement for differentiable prediction and observation models. 
Further note that this entire model is defined without making any distinction between different types of articulations.

\begin{figure}
\setlength{\fboxrule}{0.0pt}
\framebox[8.5cm]{}
\centering
\includegraphics[width=0.9\linewidth]{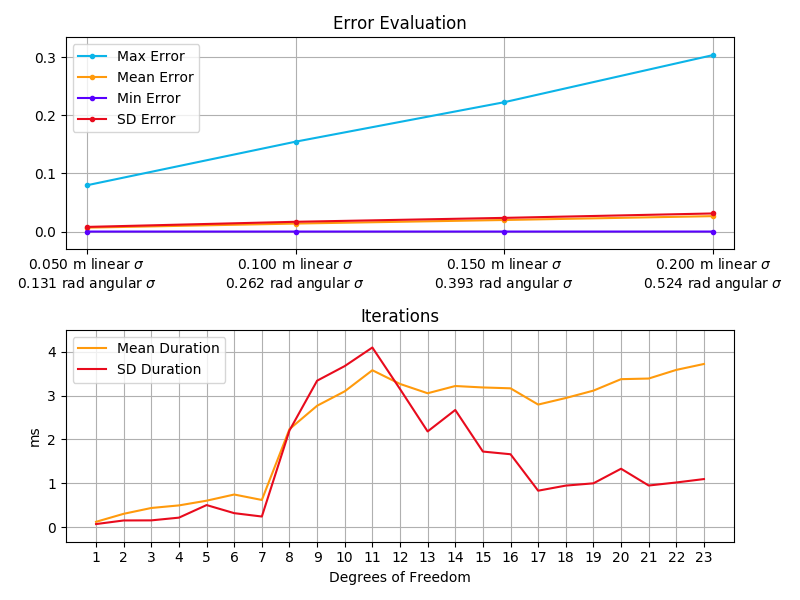}
\caption{Results of the EKF estimator evaluation. Top: The EKF estimates the state reliably within a low margin of configuration space error. Bottom: The mean duration of a prediction and update step rises with the number of DoF but remains online-capable. The stark increase in iteration duration is caused by the rotational DoF in the model. These generate larger observations and are thus more computationally expensive.}
\label{fig:ekf_results}
\end{figure}

\sectitle{Results}
We evaluate the state estimator with differing translational and rotational noise. For each of those conditions we draw 300 groundtruth poses $\hat{\vec{q}}$ and generate clean observations $\hat{\vec{z}} = h(\hat{\vec{q}})$ from these poses. Adding noise according to $\sigma_t, \sigma_r$ to $\hat{\vec{z}}$ we generate 25 observations $\vec{z}_t$ which we input into the EKF.
For each $\hat{\vec{q}}$ we initialize $\vec{q}_0 = \frac{1}{2} (\ub{\vec{q}} +  \lb{\vec{q}})$ in the center of its configuration space as indicated by the constraints of the articulated object. If one component of $\vec{q}$ is unconstrained, we default to a limit of $-\pi \leq q \leq \pi$. We initialize the covariance $\Sigma$ of the state using the square of the width of the configuration space: $\Sigma_0 = diag((\frac{1}{2} (\ub{\vec{q}} -  \lb{\vec{q}}))^2)$.
We deviate from the classic EKF scheme for the initial observation $\vec{z}_0$ by aggressively stepping towards that observation with a 10 step gradient descent. In a more non-convex problem space, this initialization would be replaced by particle filtering.

\begin{table}
\setlength{\fboxrule}{0.0pt}
\framebox[8.5cm]{}
\centering
\begin{tabular}{r|rr}
\toprule
\textbf{Opening Task} & $\varnothing$ Iteration & $\sigma$ Iteration \\ \midrule
       HSR (23 DoF)   & $3.3~ms$                & $2.2~ms$                   \\ 
       Fetch (18 DoF) & $1.2~ms$                & $0.5~ms$                   \\ 
       PR2 (49 DoF)   & $2.8~ms$                & $1.6~ms$                   \\
       \bottomrule
\end{tabular}

\begin{tabular}{r|rr}
\toprule
\textbf{Pushing Task} & $\varnothing$ Iteration & $\sigma$ Iteration \\ \midrule
       HSR (23 DoF)   & $1.9~ms$                & $0.02~ms$                   \\ 
       Fetch (18 DoF) & $2.3~ms$                & $0.01~ms$                   \\ 
       PR2 (49 DoF)   & $2.7~ms$                & $0.02~ms$                   \\
       \bottomrule
\end{tabular}
\caption{Duration of iterations for generating control signals during grasped opening of the three target objects and during pushing of objects in the IAI-kitchen using the HSR, Fetch, and PR2.}
\label{tab:push_performance}
\end{table}

\figref{fig:ekf_results} shows the results from the evaluation. First, we observe that our framework works correctly as it estimates the groundtruth pose $\hat{\vec{q}}$ closely, even under increasing noise.
\bedit We are interested in the run-time performance of our filter, as the evaluation of symbolic mathematical expressions can be quite slow. \eedit We view the filter as online-capable with a mean iteration time below $4~ms$ even when estimating the 23 DoF of the full kitchen. A spike can be observed at 8 DoF due to the setup of the experiment. The set of DoF is increased incrementally, across the experiments. The eighth DoF happens to be the first rotational joint in the setup which has larger observations $h$, increasing the size of the residual covariance and thus the computational cost of inverting it.

\subsection{Model-Agnostic Articulated Object Manipulation}

In this section, we evaluate the utility of our framework for mobile manipulation of articulated objects by formulating two model-agnostic controllers \bedit for grasped object manipulation and object pushing \eedit of which we do roll outs with different robots and objects. 
We formulate our controllers as constrained optimization problems as proposed by Fang~\etal~\cite{fang2016learning}. In this formulation, a controller is a function $f: \R^m \rightarrow \R^n$ mapping a vector of observable values to a vector of desired velocities.

\bedit In the case of both experiments, we load the robots, and objects either from a URDF file, or assemble them using custom operations in Python code. In the case of the robots, we insert an operation that connects the robot base to the world frame using an appropriate articulation. Both controllers only operate on the symbolic mathematical forward kinematics of robots, and objects and query relevant constraints from the model in the final steps of generating the actual control problem. \eedit

\subsubsection{Model-Agnostic Grasped Object Manipulation}
\label{sec:grasped_manipulation}

In this experiment, we select a manipulation task involving a static grasp. We define a goal position in an articulated object's configuration space and seek to generate a robotic motion that will move the object to this position. Using our articulation model we generate a controller generating velocity commands.

As discussed in \secref{sec:introduction}, we show that it is possible to develop model-agnostic applications using our differentiable and computationally complete model. To investigate this hypothesis, we keep our generation procedure for the controller as general as possible.
The procedure only takes the FK expression $\tf{W}{T}{E}$ of the robot's end-effector, the FK expression of a grasp pose attached to the grasped part of the object $\tf{W}{T}{G}$, a goal position $\vec{q}^o$ in the object's configuration space, and an articulation model $\A$ to query for relevant constraints.

\sectitle{Controller} Our generated controller consists of two QPs. The first driven by a proportional control term determines a desired velocity $\vec{\fd{q}}^o_{des}$ in the object's configuration space. This velocity is integrated for a fixed time step $\Delta t$, $\vec{q}^o_{t+\Delta t} = \vec{q}^o_{t} + \Delta t \vec{\fd{q}}^o_{des}$. Given this integrated position, the second QP satisfies the constraint $\app{\tf{W}{T}{E}}{\vec{q}^r_{t+\Delta t}} = \app{\tf{W}{T}{G}}{\vec{q}^o_{t+\Delta t}}$. Finally, the desired robot velocity $\vec{q}^r_{des} = \frac{\vec{q}^r_{t+\Delta t} - \vec{q}^r_{t}}{\Delta t}$ is calculated from the delta of the positions and scaled linearly to satisfy the robot's velocity constraints queried from $\A$.

\sectitle{Results} We perform roll outs of our controller manipulating different household objects using different robots in simulation. Our objects involve a simple hinged and a simple prismatic articulation but also a novel folding kinematic which cannot be modeled using existing articulation model frameworks. \figref{fig:rw_target_objects} depicts these objects. We use models of the PR2, Fetch, and HSR to solve this task. These robots vary significantly in their kinematics. The PR2 with its omnidirectional base and its $7$-DoF arms is the most articulated out of the three. While the Fetch has a $7$-DoF arm, its maneuverability is hindered by its differential drive base. The HSR's base is omnidirectional, but its arm only has $4$ DoF. We initialize all three robots with the handle of the target \emph{in-hand} and command them to open the object.\looseness=-1

The generated motions are effective though unintuitive at times. We observe that the controller produces very varied solutions for all robots, despite only having access to the \sedit{forward kinematics} of the robot and \sedit{the} object, and constraints about both queried from the model. 
We measure the time it takes to generate the next control signal to get an indication of whether this approach can be easily deployed in an online system.
Our measurements collected in \tabref{tab:push_performance} show some variance in the overall time taken to compute the next signal. From the individual measurements taken during the roll outs, we conclude this variance to be caused by the second QP iteratively satisfying the IK constraints. Nonetheless, we view the proposed controller and our approach as online capable as its update rate never drops below 100 Hz, even in the worst cases.\looseness=-1

\subsubsection{Model-Agnostic Object Pushing}
\label{sec:pushing}

\begin{figure}[t]
    \setlength{\fboxrule}{0.0pt}
    \framebox[8.5cm]{}
    \centering
    \includegraphics[width=2.8cm]{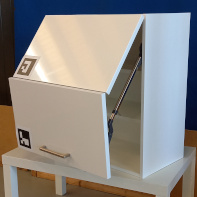}
    \includegraphics[width=2.8cm]{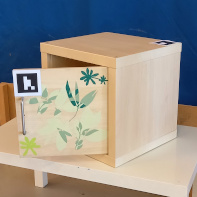}
    \includegraphics[width=2.8cm]{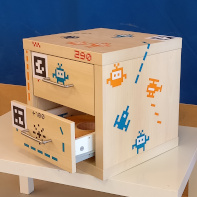}
    \caption{From left to right simulation and real-world folding door, simple hinge, and simple prismatic articulation.}
    \label{fig:rw_target_objects}
\end{figure}

In this section, we employ our framework to generate a closed-loop controller to push an articulated object towards a desired configuration space goal. As in the previous section, we keep the controller definition agnostic of both the robot's and articulated object's kinematics.
We use this controller formulation to generate motions for various robots manipulating articulated objects in a kitchen environment.

\begin{figure}
\setlength{\fboxrule}{0.0pt}
\framebox[8.5cm]{}
\centering
\includegraphics[height=3.4cm]{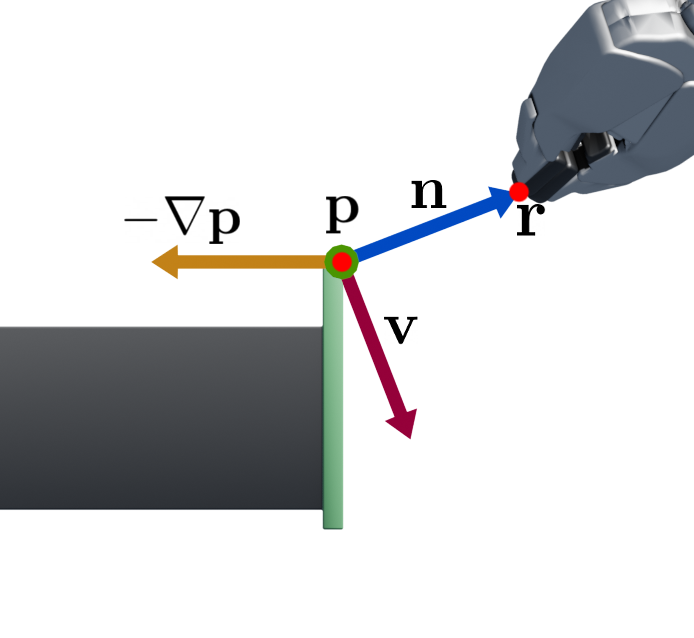}
\includegraphics[height=3.4cm]{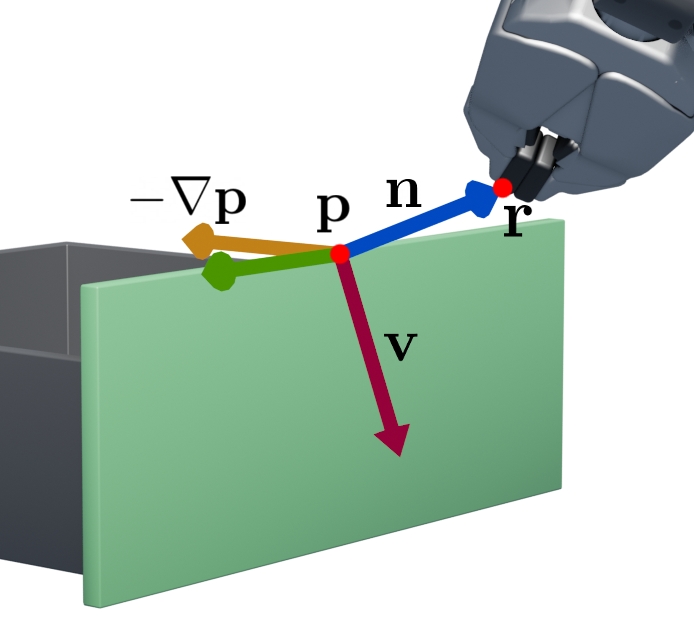}
\caption{Points and vectors used to generate motions for a push. $\vec{r}$ (red): Contact point on the robot; $\vec{p}$ (red): Contact point on the object (the front of a drawer in this case); $\vec{n}$ (blue): Contact normal; $-\nabla \vec{p}$ (yellow): Negative gradient of $\vec{p}$; Green arrow: neutral tangent; $\vec{v}$ (dark red): Active tangent.}
\label{fig:push_vectors}
\end{figure}

\sectitle{Controller}
Given an articulated object $o$ that should be pushed by the robot using link $l$, we use our \sedit{implementation's} integration with a collision checking library to obtain points $\vec{p}, \vec{r}$ that are the closest points on the object and robot and additionally the \sedit{contact} normal $\vec{n}$ pointing from $\vec{p}$ towards $\vec{r}$.
We use these entities to define constraints that enforce plausibility in the generated pushing motion, \ie robot and object have to touch ($\norm{\vec{p} - \vec{r}} = 0$) and the object can only be pushed not pulled ($\vec{\fd{p}}^T\vec{n} < 0$).
In addition to these constraints, we define an additional navigation heuristic that navigates the robot's contact point $\vec{r}$ to a suitable location where it can push $\vec{p}$ in the desired direction. This is achieved by combining the gradient vector $\nabla \vec{p}$ with $\vec{n}$ to derive an active navigation direction $\vec{v}$ in which the angle between the normal and the gradient widens. \figref{fig:push_vectors} visualizes this concept.
Finally, the controller avoids contact with all other parts of the object that will move during this interaction by querying these objects from the model $\A$ on the basis of $\var{\vec{p}}$.

\sectitle{Results}
We evaluate our controller in a simulated environment with three different robots. The environment contains the IAI-kitchen. The task of the robot is to close its doors and drawers. We select $9$ drawers and $3$ doors with different axes of rotation for our interactions.
The doors and drawers are opened $40 cm$ or $0.4 rad$ and the interactions are executed individually with the robots starting at the same position for each interaction. 
The link $l$ with which a robot should push and the object $o$ that should be pushed are given for each interaction.
As a particularly challenging object, we also confront the controller by closing the cupboard with a folding door as depicted in \figref{fig:rw_target_objects}. 
Again, we roll out the trajectories generated by the controller described above. As with the previous task, we again use models of the PR2, Fetch, and HSR to solve this task due to their kinematic diversity.

The generated motions are again effective though some appear more stable than others. Especially the differential drive on the fetch produces very far-reaching motions when trying to operate objects which are far away from the initial location.
In the case of the folding door, the generated motions seem possible but optimistic. Note that it is necessary to be able to exert force onto different parts of the handle during different states of closing the door. Initially, the handle has to be pushed from the top and later from the front. Especially the start of the motion seems to be challenging for the controller using the Fetch and HSR. We discovered this to be a fault caused by the geometry of the robots' fingers. Nevertheless, we view the overall concept to be a success, especially given the fast control cycle as shown in \tabref{tab:push_performance}.

\subsection{Real-World Object Manipulation}

After demonstrating how our proposed framework can be used in different individual robotic problems, we deploy our framework on a real robotic system. We use the model-agnostic manipulation controllers from the previous section with a state estimator and deploy them to interact with the objects in \figref{fig:rw_target_objects}.

\sectitle{Setup}
\bedit As in the previous experiments, we load the models for robots and objects either from URDF or build them using our operations interface. \eedit
The 6D pose observations of known frames in the articulation model of the handled object are collected using AR-markers. A model-agnostic state estimator uses these observations to estimate the current object \sedit{configuration space} pose. This pose is processed by a simple behavior, wrapping the grasped manipulation controller from \secref{sec:grasped_manipulation}. The behavior monitors a subset of the object's pose and attempts to \emph{open} or \emph{close} all parts of an object. Once a target is selected, a simple grasping routine is executed, the object is moved to the desired position, and finally released. To close an object, the pushing controller from \secref{sec:pushing} is executed directly until the object is observed as closed.  We perform the evaluation on a PR2 robot with a whole-body velocity control interface\bedit, including the base, \eedit and run each task three times in a row as a minimal measure of system stability.

\begin{figure}[t]
\setlength{\fboxrule}{0.0pt}
\framebox[8.5cm]{}
    \centering
    \includegraphics[width=8cm]{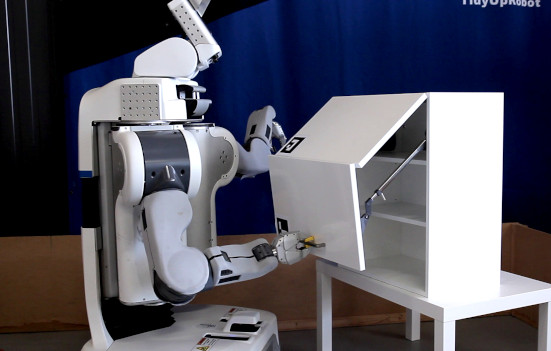}
    \caption{PR2 operating real folding door kinematic during our experiments.}
    \label{fig:rw_execution}
\end{figure}

\sectitle{Results}
We observe that the system is able to achieve the task with all three objects. \figref{fig:rw_execution} shows the PR2 interacting with the folding door. Although the execution is successful, it is not entirely efficient. The controller does not consider the dynamics of the robot, which causes a discrepancy in prediction and execution when moving the entire robot. This discrepancy leads to a very slow motion execution, especially with the spring-loaded folding door. \bedit The lack of a dynamics model is also betrayed in the final moments of the door closing when the robot has pushed the door past the point at which it can be held open by its springs. Here the robot's model is not expecting the door to fall shut automatically, which is why it chases the handle for a few moments before the state estimation recognizes the door as shut. \eedit
This is to be expected as we did not model the dynamics\bedit, but the experiments have demonstrated vividly that dynamics modeling, and more physically accurate state predictions are of prime importance for the future usage of our approach in real robotic manipulation\eedit. We view the overall executions as a successful demonstration of how our approach can be utilized throughout a live robotic system. 

\section{Conclusions}

In this paper, we proposed a novel framework for modeling articulated \bedit structures\eedit. With our approach, we sought to overcome the computational opaqueness of currently existing frameworks, which we view as limiting the inclusion of new articulations in these frameworks.
We contributed a theoretical description of our framework using computationally transparent symbolic mathematical expressions in building such models.
To address our second identified challenge, the rigidity of articulation models in live robotic systems, we \bedit introduced our reference implementation including mechanisms for exchanging our models among components of a live robotic system. \eedit
In our evaluation, we demonstrated our approach to provide greater modeling capabilities and investigated whether \emph{model-agnostic} robotic skills are possible with our approach by implementing and evaluating such skills. Finally, we demonstrated the suitability of our approach to real-world mobile robotics by deploying the aforementioned skills in a live robotic system.

\bedit For future work, we see a need to investigate how our models connect with existing approaches in model estimation. In our evaluation, we provided the models of robots and objects manually, however, these should be estimated using approaches similar to the ones we introduced in \secref{sec:related_work}. Further, our real-world experiments have shown that the modeling of the dynamics of articulated structures requires attention. \eedit




\balance
\bibliographystyle{IEEEtran}
\bibliography{main}

\end{document}